\documentclass[times,twocolumn,final,authoryear]{elsarticle}

\usepackage{prletters}
\usepackage{framed,multirow}

\usepackage{amssymb}
\usepackage{latexsym}
\usepackage{tikz}
\usetikzlibrary{shapes,arrows}
\usepackage{amsmath}
\usepackage{amsfonts}
\usepackage{bbm}
\usepackage{caption}

\usepackage{url}
\usepackage{xcolor}
\definecolor{newcolor}{rgb}{.8,.349,.1}

\graphicspath{{Images/}}

\setcitestyle{numbers,open={((},close={))}}

\begin{document}
\thispagestyle{empty}
                                                             
\setcitestyle{square}

\clearpage
\thispagestyle{empty}
\ifpreprint
  \vspace*{-1pc}
\fi

\clearpage
\thispagestyle{empty}

\ifpreprint
  \vspace*{-1pc}
\else
\fi

\clearpage

\ifpreprint
  \setcounter{page}{1}
\else
  \setcounter{page}{1}
\fi

\begin{frontmatter}

\title{An improved Bayesian TRIE based model for SMS text normalization}

\author[1]{Abhinava Sikdar\corref{cor1}} 
\cortext[cor1]{Corresponding author: 
  Tel.: +91-8700572115}
\ead{mt1170724@iitd.ac.in}
\author[2]{Niladri Chatterjee}

\address[1]{Department of Mathematics, IIT Delhi, New Delhi, India}
\address[2]{Soumitra Dutta Chair Professor in Artificial Intelligence, Department of Mathematics, IIT Delhi, New Delhi, India}

\received{1 May 2013}
\finalform{10 May 2013}
\accepted{13 May 2013}
\availableonline{15 May 2013}
\communicated{S. Sarkar}

\begin{abstract}
Normalization of SMS text, commonly known as texting language, is being pursued for more than a decade. A probabilistic approach based on the Trie data structure was proposed in literature which was found to be better performing than HMM based approaches proposed earlier in predicting the correct alternative for an out-of-lexicon word. However, success of the Trie-based approach depends largely on how correctly the underlying probabilities of word occurrences are estimated. In this work we propose a structural modification to the existing Trie-based model  along with a novel training algorithm and  probability generation scheme.  We prove two theorems on statistical properties of the proposed Trie and use them to claim that is an unbiased and consistent estimator of the occurrence probabilities of the words. We further fuse our model into the paradigm of noisy channel based error correction and provide a heuristic to go beyond a Damerau–Levenshtein distance of one.\\
\textbf{\textit{Keywords:} SMS text normalization, Noisy channel, Trie, Theory of estimation}
\end{abstract}

\begin{keyword}
\KWD Keyword1\sep Keyword2\sep Keyword3

\end{keyword}

\end{frontmatter}


\section{Introduction} \label{Section1}
SMS text normalization focuses on translating texting language, often filled with abbreviations and marred by typing errors into plain English text. With the smartphone revolution and massive popularisation of social media platforms, users often transmit messages consisting of up to thousands of words in a single day. However, these text messages consist of numerous abbreviations and errors. This often arises due to a lack of formalities between users, human error, and in more severe cases due to disabilities. With an increase in the screen sizes, this is becoming more of a concern especially when the user resorts to one handed typing. Thus, shorter message length and semantic unambiguity end up working antagonistically which gives shape to a compressed, non-standard form of language called NetSpeak \cite{crystal_2006}, commonly known as the texting language. Unfortunately, traditional NLP methods perform rather poorly when dealing with these kinds of texts \cite{10.5555/1214993}. As described in \cite{Kobus2008NormalizingSA}, texting language contains many non-standard abbreviations, have unique characteristics and behave quite differently which may be the reason for poor performance of traditional NLP methods.
\newpage
In \cite{Li2012NormalizationOT} the problem was approached by leveraging the phonetic information but required a dictionary for phonetic transcription. \cite{article1} used a dictionary for erroneous words but used an ambiguous ranking mechanism. \cite{6495103} used a completely heuristic rule based model and employed a dictionary for typos as well which in all the three cases caused the time complexity to grow with the size of the dictionary. \cite{Veliz2019ComparingMA} compared machine translation models such as SMT and NMT and showed their critical dependence on high quality training data. In \cite{article}, a Hidden Markov Model based approach was used for each word present in the corpus to model all possible normalized or corrected texts and their probabilities of occurrence. This was further used to rank all the possible normalised words and the highest ranking word would be selected for output as the semantically correct word. In \cite{inbook}, a Trie based probability generation model was proposed, and was shown to outperform HMM-based models whenever the incorrect word was within an edit distance one after prepossessing for phonetic errors. However, in some cases the target word did not end up having the highest rank. For example, `mate', the intended target word was ranked $4^{th}$ in the suggestions list for the word `m8'.\\
Through this work we address the limitations of this Trie based probability generation model. We make a set of improvements  over the scheme proposed in \cite{inbook} which we consider to be the baseline system for the present work. Firstly, unlike \cite{inbook} where a new word is added in a Trie only once when it is encountered for the first time during the training phase,  a dynamic training environment is used in the proposed model. In the proposed model the Trie is initially constructed with a certain corpus of most commonly used words and is then deployed. After deployment, the tries learns dynamically  from the texting habits of the users over time i.e. it keeps adding the same words repeatedly as and when they are encountered. We show that the new model in the described environment is able to estimate the real probability of occurrence of a word through its own new probability generating algorithm. Hence it facilitates 0-1 loss minimisation in a decision theoretic setup. This in turn minimizes the chances of the target word not being ranked as the highest one. \\
Additionally, as explained in \cite{Keyboards2019ImprovingTA} , it is computationally infeasible to generate all the words which are at
an edit distance two from a given word. This limited the spelling correction abilities of the baseline model. In this work we propose some novel heuristics that help in overcoming the above shortcoming significantly.
\\
The contributions of the present work are
\begin{enumerate}
    \item We suggest a structural modification to the Trie-based reasoning scheme to improve model accuracy and performance.
    \item We prove mathematically that in the described dynamic environment, the expectation of the thus generated Trie probability equals the occurrence probability for each word.
    \item We further prove that the Trie probability of each word in the corpus almost surely(a.s.) converges to its occurrence probability.
    \item Develop a set of heuristics for error correction.
    \item Provide empirical results through simulations to support the presented theorems and highlight the superiority of the new model.
\end{enumerate}

\section{Background} \label{Section2}
From existing literature on the behaviour of users while using texting language \cite{Grinter2001}, one can classify most of the linguistic errors as:
\begin{enumerate}
    \item Insertion Error: Insertion of an extra alphabet in the word. Eg. `hoarding'$\rightarrow$`hoardingh'
    \item Deletion Error: Deletion of an alphabet from the word. Eg. `mobile'$\rightarrow$`moble'
    \item Replacement Error: Replacement of an alphabet by another in the word. Eg. `amazing'$\rightarrow$`anazing'
    \item Swap Error: Swapping of two adjacent characters in the word. Eg. `care'$\rightarrow$`caer'
    \item Phonetic Error: A group of alphabets is replaced by another group of alphabets or numbers having the same sound. Eg. `tomorrow'$\rightarrow$`2morrow'
\end{enumerate}
To deal with these errors, an elaborate error correction algorithm has already been proposed in the baseline model wherein given an erroneous/non-lexical word, a suggestion list of possible target words was prepared. The probability of occurrence for each word in the list was generated using a Trie based algorithm as described in Section \ref{Section2.2} which was used for ranking the words in the suggestion list. The highest ranking word was given out as the likely intended word.

\subsection{Design of the TRIE}
Trie is a memory efficient data structure primarily used for storing of words. The search complexity in the data structure is $\mathcal{O}$($M$), where $M$ represents the number of characters in the longest word stored in the Trie which makes it a computationally efficient data structure for the problem.\\
 Each node contains an integer variable \textit{Count} and a Boolean variable \textit{End-Of-Word}. The Trie is set up by initializing it with a number of English words. \textit{Count} for each node is initiated to zero and incremented by one each time a word is inserted through that node. \textit{End-Of-Word} represents if a word ends at the given node and is set to \textit{True} at the ending node of each word. Each time a new word is inserted, the \textit{Count} variables of the passing nodes are updated and if at any point the next node is not already present for the insertion of a character, a new node is created and added as a child. At the end of the new word, \textit{End-Of-Word} is switched from \textit{False} to \textit{True}.
\begin{figure}[t]
    \centering
    \includegraphics[width=\linewidth, height=9cm]{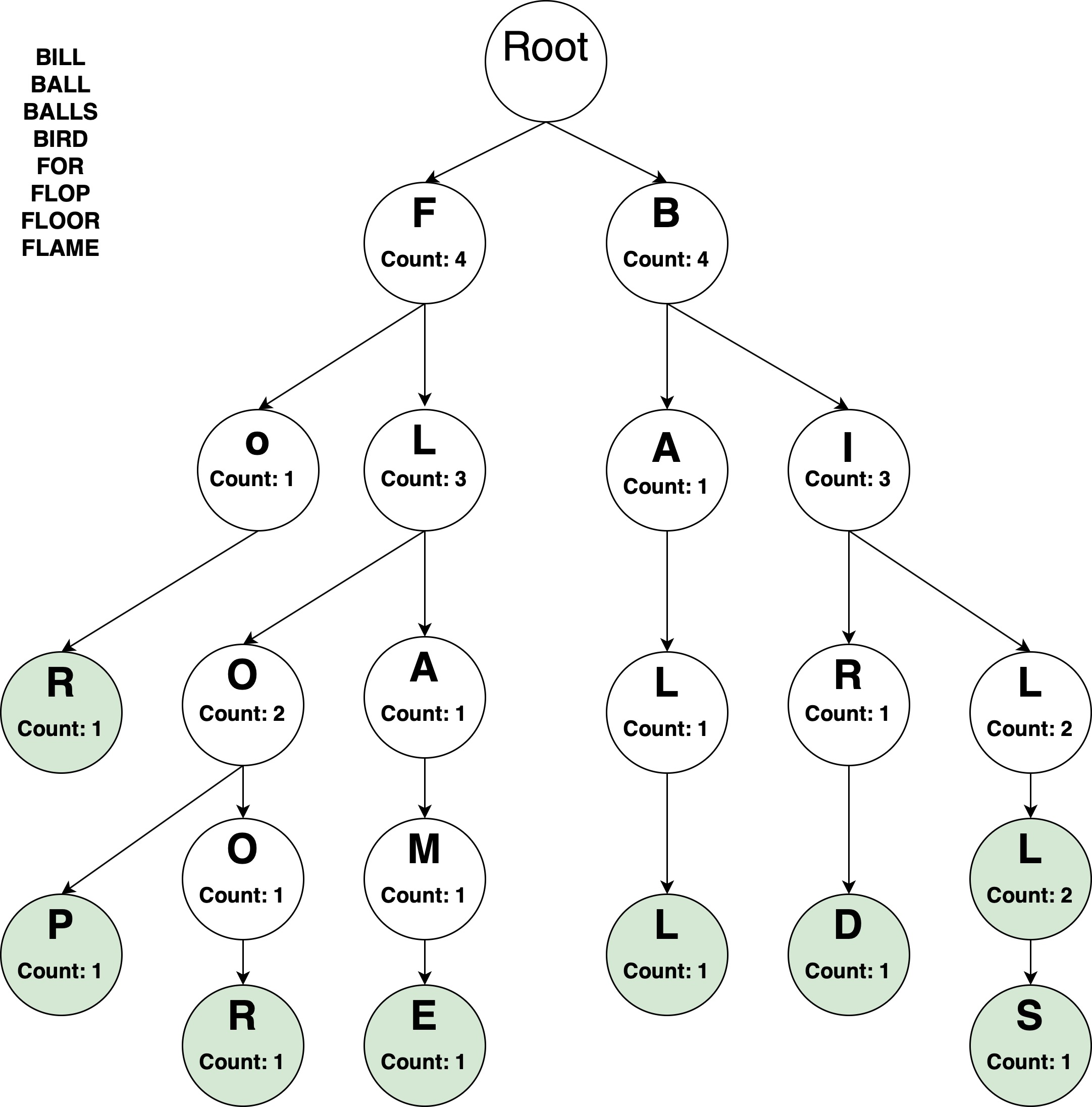}
    \caption{Trie: The model has been trained with 8 words each once. Green represents a \textit{True} value for \textit{End-Of-Word} for the coloured node.}
    \label{fig:Trie}
\end{figure}

\subsection{TRIE probability of a Word}\label{Section2.2}
Assigning probability of occurrence to a word through the Trie is a task of prime importance in the model. In the baseline model this was done in a simplistic manner as described below.  \\
The probability of choosing the $i^{th}$ child node was estimated by dividing its \textit{Count} variable, to be referred as \textit{Count}(\textit{i}) by \textit{Total}(\textit{i}) which was defined as:
\begin{equation}\label{TotalEqn}
    Total(i)=Count(i)+\sum_{j\in Sibling(i)}Count(j) 
\end{equation}
Where \textit{Sibling}(\textit{i}) refers to the set of all the other child nodes of the common parent node. Hence the probability of the $i^{th}$ child node was set to be:
\begin{equation} \label{TrieProbabEqn}
    P(i)=\frac{Count(i)}{Total(i)}
\end{equation}
Further, to get the probability of a string $s_1s_2\ldots s_n$ from the Trie, the conditional rule of probability was used as:
\begin{equation}
    P(s_1s_2\ldots s_n)=P(s_1)P(s_2|s_1)\ldots P( s_n|s_1s_2\ldots s_{n-1})\nonumber
\end{equation}
For example, consider the string 'bird', then
\begin{equation}
    P(bird)=P(b)P(i|b)P(r|bi)P(d|bir) \nonumber
\end{equation}
Hence, from Fig.\ref{fig:Trie} we get that $P(bird)=\frac{1}{2}\times\frac{3}{4} \times\frac{1}{3} \times\frac{1}{1}=\frac{1}{8}$.
\section{The Improved Trie}\label{Section3}
We consider the user to be modelled as a corpus of words denoted by $\mathcal{S}$ where each word $w_i\in \mathcal{S}$ has an occurrence probability of $p_i$. Note that then the training and probability calculations as proposed in the baseline Trie does not account for an important case. This is when the corpus of words $\mathcal{S}$ contains two words say $w_1$ and $w_2$ with occurrence probabilities $p_1$ and $p_2$, $p_1\neq p_2$ such that $w_1$ is a prefix of $w_2$ or vice versa. For example consider the two words 'bill' and 'bills' in Fig.\ref{fig:Trie}. No matter what the occurrence probabilities of those two words are, they will always end up with the same Trie probability\footnote{Generally, in more branched Tries, whenever $w_1$ is a prefix of $w_2$, Trie probability of $w_2$ $\leq$ Trie probability of $w_1$} even after training the Trie a large number of times after random sampling from $\mathcal{S}$. While the deviation from occurrence probabilities might not vary a lot in the example considered since one word happens to be the plural of the other, it will matter much more in words such as 'lips' \& 'lipstick' and `dear' \& `dearth'.\\
To overcome this we propose the introduction of a dummy node. This dummy node is added as a child of every node which has \textit{End-Of-Word} set as \textit{True} but has at least one non-dummy node as its child. The \textit{Count} variable of this dummy node is set to be the count variable of the parent reduced by the count variable of all the siblings of the new dummy node.
\begin{equation} \label{dummyEqn}
    Count(D)=Count(P)-\sum_{j\in Siblings(D)}Count(j)
\end{equation}
\\
Where $D$ denotes the dummy node and $P$ denotes the parent of the dummy node. Algorithm I in this section outlines the procedure for training the new Trie as described above.
\\
In addition to the training algorithm, we also propose Algorithm II which is used to generate the Trie probability of a given string. \\
Further, to justify usage of the new Trie mathematically, the proposed training algorithm and the algorithm for the Trie probability generation, we present Theorem 3.1, Theorem 3.2 and Corollary 3.3.
\newpage
\hrule
\vspace{0.03cm}
\textbf{Algorithm I: The Training}
\hrule
\vspace{0.1cm}
\textbf{Input:} $w$, root\\
\textbf{Initialization:} $str\leftarrow w$, $m\leftarrow str$.length, $parent \leftarrow$ root, $Flag\leftarrow 0$\\
\textbf{For} $j\leftarrow 1$ to $m$
\begin{enumerate}
    \item \textbf{If} $parent$.children does not contain $str$.charAt[$j$] node
    \begin{enumerate}
        \item Insert new node corresponding to $str$.charAt[$j$]
        \item $Flag\leftarrow 1$
    \end{enumerate}
    \item $parent\leftarrow$ child corresponding to $str$.charAt[$j$]
    \item $parent$.Counter $\leftarrow parent$.Counter$+1$
\end{enumerate}
$parent$.End-Of-Word$\leftarrow$True\\
\textbf{If} a dummy node $D$ is already a child of $parent$
\begin{enumerate}
    \item Increment Counter of $D$ by 1
\end{enumerate}
\textbf{Else if} $Flag$ is 0 and $parent$ has at least 1 non-dummy child
\begin{enumerate}
    \item Insert new dummy node as a child of $parent$
    \item set Counter of dummy as in Eqn.(\ref{dummyEqn})
\end{enumerate}
 \textbf{End}
\vspace{0.1cm}
\hrule
\vspace{0.15cm}
\hrule
\vspace{0.08cm}
\textbf{Algorithm II: The Trie Probability Generation}
\hrule
\vspace{0.1cm}
\textbf{Input:} $w$, root\\
\textbf{Output:} Trie probability of $w$\\
\textbf{Initialization:} $str\leftarrow w$, $m\leftarrow str$.length, $parent \leftarrow$ root, $probab\leftarrow 1$\\
\textbf{For }$j\leftarrow1$ to $m$
\begin{enumerate}
    \item \textbf{If} $parent$.children does not contain str.charAt[j] node
    \begin{enumerate}
        \item \textbf{return} $0$
    \end{enumerate}
    \item $child\leftarrow$ child corresponding to str.charAt[j]
    \item Update $probab$ using Eqn.(\ref{TotalEqn}) \& Eqn.(\ref{TrieProbabEqn}) for $child$
    \item $parent\leftarrow child$
\end{enumerate}
\textbf{If} $parent$.End-Of-Word is True
\begin{enumerate}
    \item \textbf{If} $parent$ has dummy node
    \begin{enumerate}
        \item Update $probab$ using Eqn.(\ref{TotalEqn}) \& Eqn.(\ref{TrieProbabEqn}) for $parent$
    \end{enumerate}
    \item \textbf{return} $probab$
\end{enumerate}
\textbf{return} $0$\\
\textbf{End}
\vspace{0.1cm}
\hrule
\vspace{0.3cm}

\begin{figure}[t]
    \centering
    \includegraphics[width=\linewidth, height=8cm]{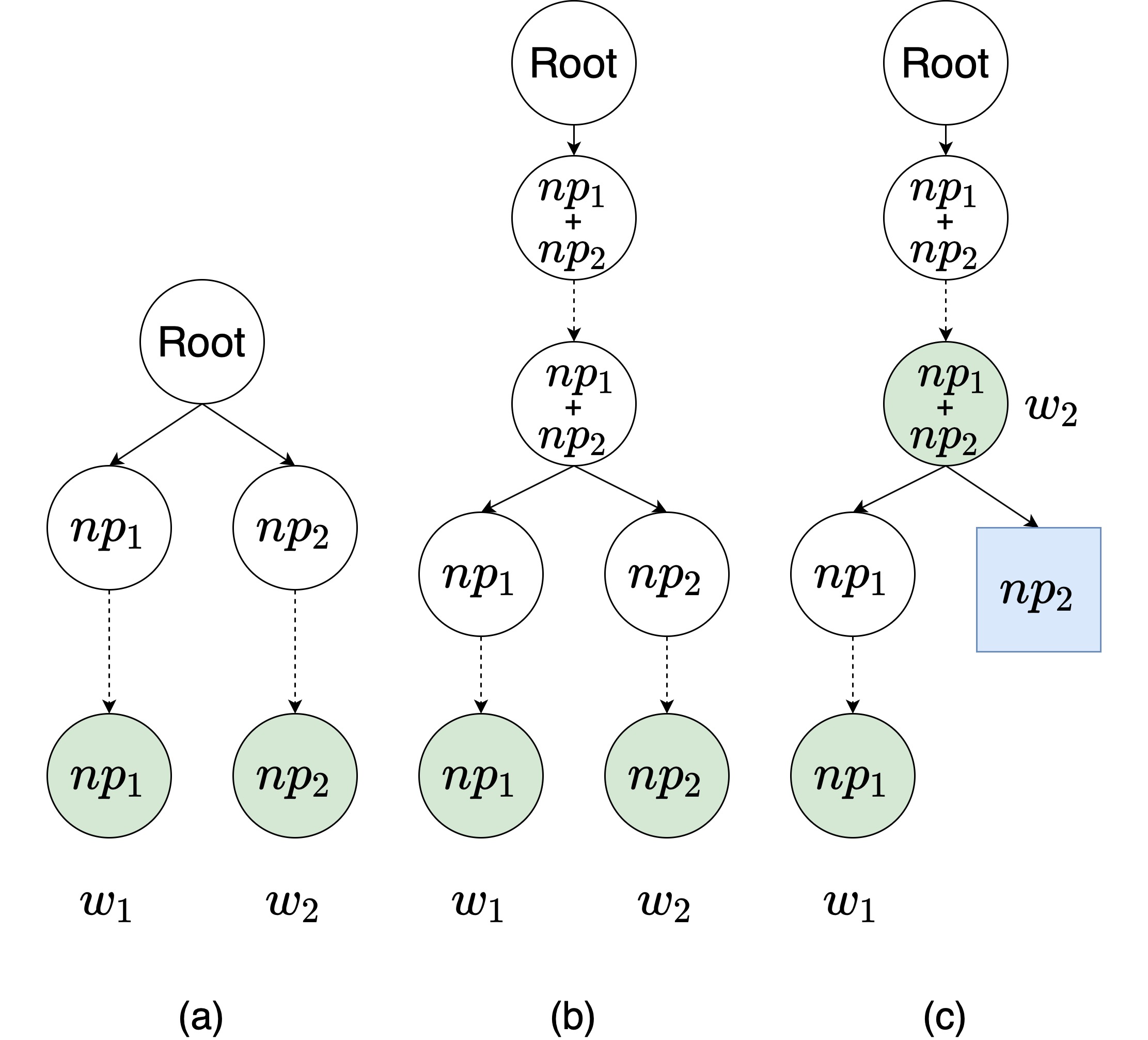}
    \caption{Induction Base Cases}
    \label{fig:basecaseDiag}
\end{figure}

\textbf{Theorem 3.1:} Let $\mathcal{S}$ denote a corpus of finite words. Let $w_i\in \mathcal{S}$ be a word of the corpus with an occurrence probability $p_i$ and let $\hat{p}_i$ denote the word probability generated by the Trie. Let the Trie be trained $n$ number of times after randomly sampling(with replacement) from $S$ such that the Trie is trained with each $w_i\in \mathcal{S}$ at least once. Then, $\mathbb{E}[\hat{p}_i] = p_i$.
\\ \\
\textit{Proof:} We use strong induction over the number of words present in the corpus to prove the theorem.\\
\textbf{Base Case:} Consider a corpus $\mathcal{S}$ consisting of two words $w_1$ and $w_2$ with occurrence probabilities $p_1$ and $p_2$. Then three cases are possible as shown in Fig.\ref{fig:basecaseDiag}.\\
\textit{Case I:} The starting characters of $w_1$ and $w_2$ are different. Then, after using the random sample from $\mathcal{S}$ to train the Trie $n$ number of times. The Trie along with all the expected values of the \textit{Count} variables are shown in Fig.\ref{fig:basecaseDiag}(a). Clearly,
\begin{equation} \label{BaseCaseEqn1}
    \mathbb{E}[\hat{p}_1]=\frac{np_1}{np_1+np_2}=p_1
\end{equation}
and
\begin{equation} \label{BaseCaseEqn2}
    \mathbb{E}[\hat{p}_2]=\frac{np_2}{np_1+np_2}=p_2
\end{equation}
\textit{Case II:} Both $w_1$ and $w_2$ have a common string of characters in the front. For illustration, `more' and `most' have `mo' common at the front. Hence similar to Case I, it can be argued from Fig.\ref{fig:basecaseDiag}(b). that Eqn.(\ref{BaseCaseEqn1}) and Eqn.(\ref{BaseCaseEqn2}) hold true.\\
\textit{Case III:} $w_1$ can be obtained by appending a string of characters at the end of $w_2$ or vice versa. Here, as described earlier we introduce a dummy node, seen as the blue box in Fig.\ref{fig:basecaseDiag}(c). Similar to previous arguments, it can be seen that Eqn.(\ref{BaseCaseEqn1}) and Eqn.(\ref{BaseCaseEqn2}) hold.\\
Hence, for the base case when $|\mathcal{S}|=2$, the theorem is true.
\\
\textbf{Induction:} Assume that the theorem holds whenever $|\mathcal{S}|\leq k$ for some positive integer $k$. We now show that the theorem holds true for any corpus of size $k+1$. Hence assume that we are given the words $w_1,w_2,\ldots ,w_{k+1}$ with occurence probabilities $p_1,p_2,\ldots p_{k+1},$ respectively.

\begin{figure}
    \centering
    \includegraphics[width=\linewidth, height=8cm]{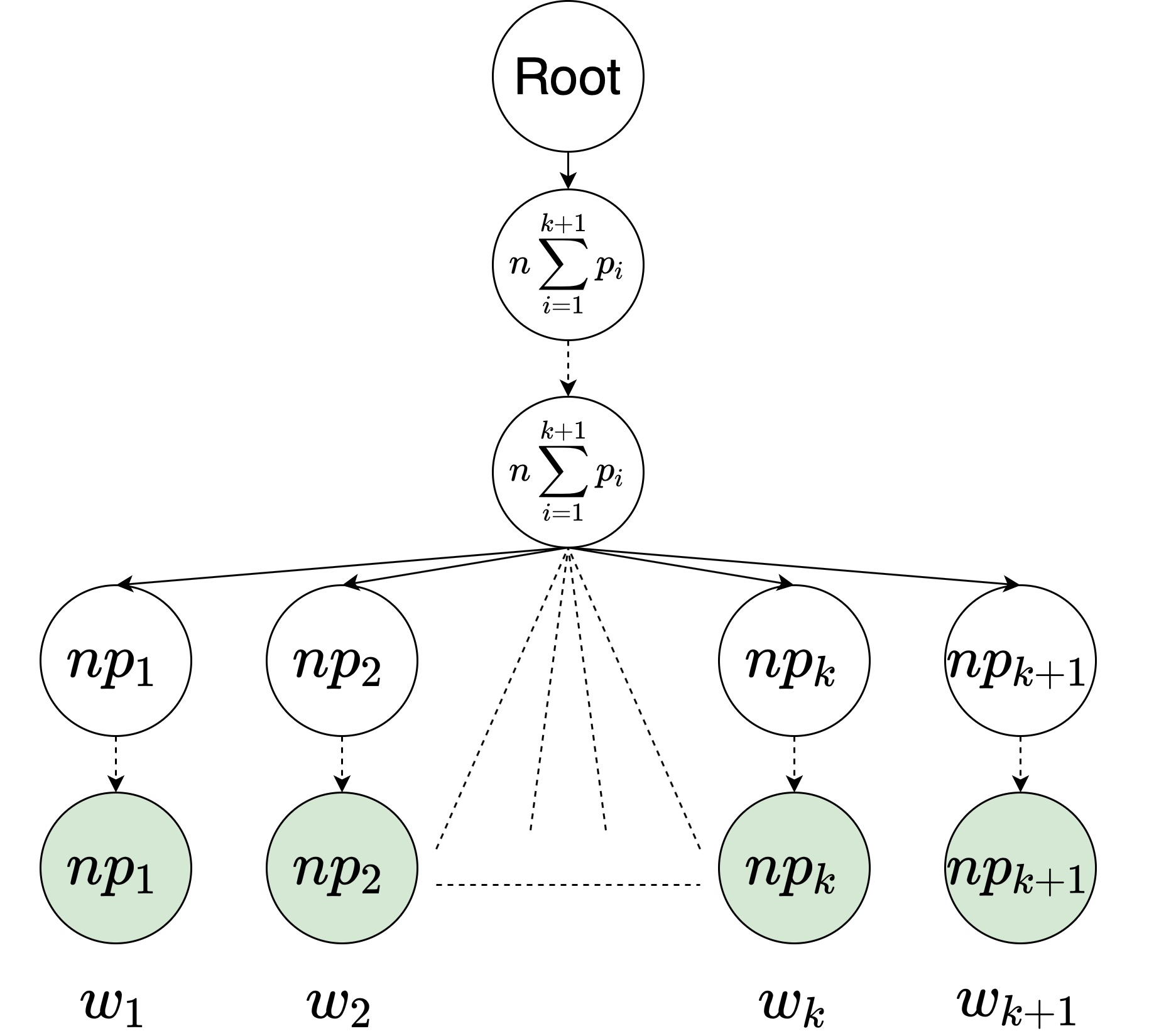}
    \caption{Induction Step Case I}
    \label{fig:inductionCase1}
\end{figure}

\begin{figure}[t]
    \centering
    \includegraphics[width=6cm, height=6cm]{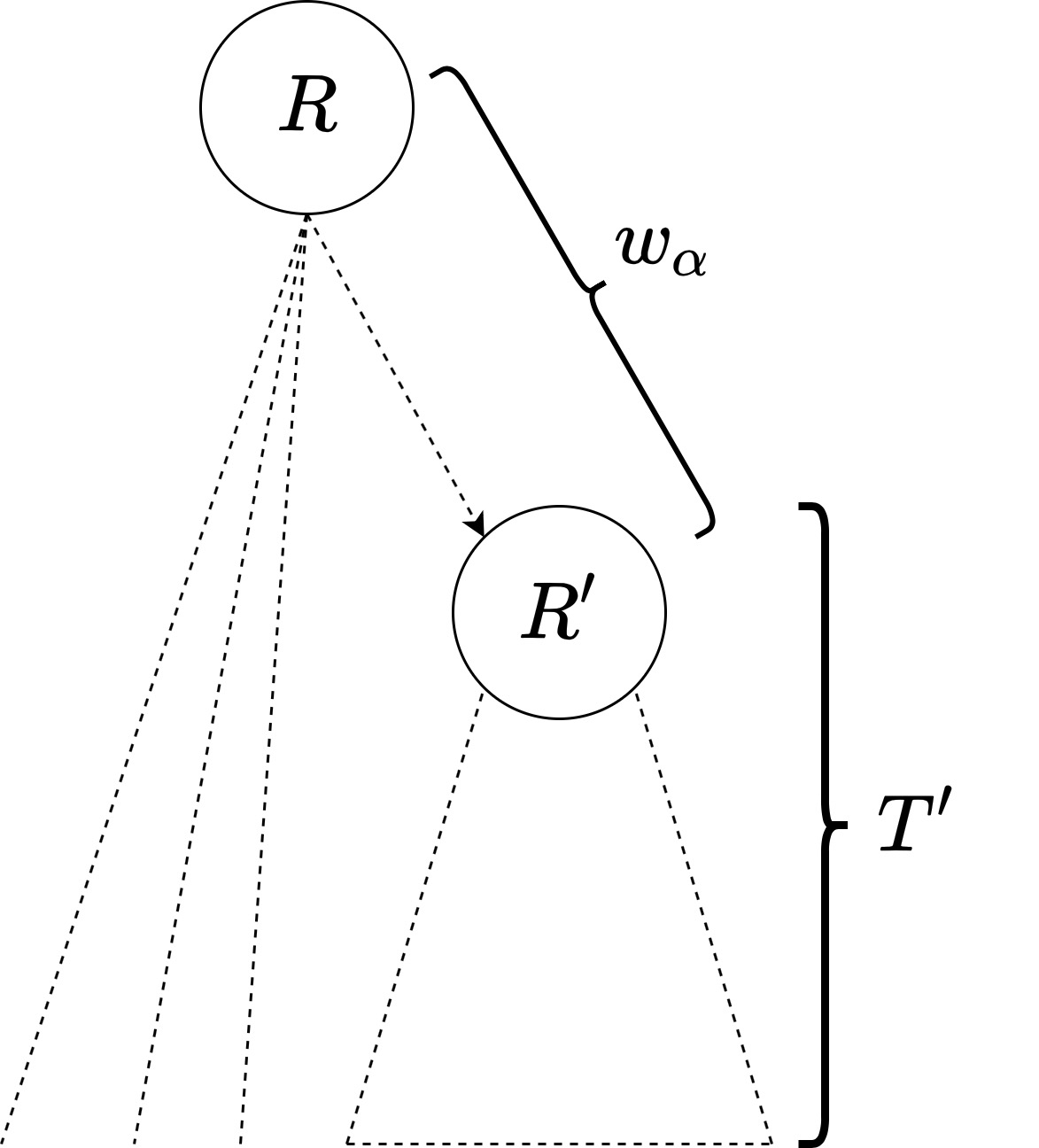}
    \caption{Induction Step Case II}
    \label{fig:inductionCase2}
\end{figure}

\textit{Case I:} Similar to Case I and Case II of the base case, all the words branch out at the same node together as in Fig.\ref{fig:inductionCase1}. The figure also shows the expected values of the $Counter$ variables after training the Trie $n$ number of times.
Hence we can easily conclude that for each $i=1,2,\ldots k,k+1$,
\begin{equation}\label{inductionEqn1}
    \mathbb{E}[\hat{p}_i]=\frac{np_i}{np_1+\ldots +np_{k+1}}=p_i
    \vspace{-0.1cm}
\end{equation}
since $\sum_{i=1}^{k+1}{p_i}=1$.\\
\textit{Case II:} In this case, consider all such possible Tries which are not covered by Case I. These Tries are in a way ``more branched" than the ones considered in Case I. Consider $T$ to be any such Trie and denote its root by $R$. Notice that there must exist a subtree $T'$, which contains strictly less than $k+1$ and at least two nodes which have their \textit{End-Of-Word} set as \textit{True}. It is easy to see that if not so, then $T$ lies in Case I which is a contradiction. Let us denote the root of $T'$ by $R'$. Let us assume that the nodes in $T'$ for which $End-Of-Word$ is set to $True$ represent the words $w_{(1)},w_{(2)},\ldots ,w_{(r)}$ with probabilities $p_{(1)},p_{(2)},\ldots ,p_{(r)}$ respectively where $2\leq r <k+1$. Now suppose that $T$ instead of having the words $w_{(1)},w_{(2)},\ldots ,w_{(r)}$, has a word whose characters are defined by travelling from $R$ to $R'$ in the Trie as shown in Fig.\ref{fig:inductionCase2}. Let this prefix word be $w_\alpha$ with a probability of $p_\alpha=p_{(1)}+p_{(2)}+\ldots +p_{(r)}$. Hence $T$ now practically contains strictly less than $k+1$ words. Hence by the induction hypothesis, when we train $T$, such that each one of the $k+2-r$ words is used for training at least once, the expectation of the Trie probabilities matches the occurrence probabilities. This means for each $w_i\in T$ whose last node doesn't lie in $T'$, Eqn.(\ref{inductionEqn1}) holds true.
Also, at the end of the training,
\begin{align}
    \mathbb{E}[Count(R')] &= n\times (p_{(1)}+p_{(2)}+\ldots+p_{(r)}) \nonumber \\
    &= \alpha \times \sum_{i=1}^{r}{\frac{p_{(i)}}{p_{(1)}+p_{(2)}+\ldots+p_{(r)}}} \label{inductionEqn2}
\end{align}
where $\alpha=n\times (p_{(1)}+p_{(2)}+\ldots+p_{(r)})$. We can now consider $T'$ to be a standalone Trie by itself, consisting of  words $w_{(1)},w_{(2)},\ldots ,w_{(r)}$ but truncated from the front so as to remove the characters of $w_\alpha$. Let these truncated words in $T'$ be denoted by $w'_{(1)},w'_{(2)},\ldots ,w'_{(r)}$ with occurrence probabilities $p'_{(1)}+p'_{(2)}+\ldots+p'_{(r)}$. Now notice that $T'$ has been trained $\alpha$ number of times and since $r<k+1$, the expectation of probabilities of $T'$ will be the same as occurrence probabilities of the truncated words which from $T$ are given by:
\begin{equation}\label{newProbabEqn}
    p'_{(i)}=P_{occurrence}(w_{i}|w_\alpha)=\frac{p_{(i)}}{p_{(1)}+p_{(2)}+\ldots+p_{(r)}}
\end{equation}

Hence, as described earlier, using induction, for each $w'_{(i)}$, $i=1,\ldots ,r$
\begin{equation}\label{InductionEqn3}
    \mathbb{E}[\hat{p}'_{(i)}]=\frac{p_{(i)}}{p_{(1)}+p_{(2)}+\ldots+p_{(r)}}
\end{equation} 
Let us consider the entire tree $T$ with its original words $w_{(1)},w_{(2)},\ldots ,w_{(r)}$ and not $w_\alpha$. Note that for each $w_{(i)}$, $i=1,2,\ldots,r$
\begin{equation}\label{InductionEqn4}
    \hat{p}_{(i)}=\hat{p}'_{(i)} \times \hat{p}_\alpha
\end{equation}
One may further notice that this is the same as $P_{Trie}(w_{(i)})=P_{Trie}(w_{(i)}|w_\alpha)\times P_{Trie}(w_\alpha)$, which implies that $\mathbb{E}[P_{Trie}(w_{(i)})]=\mathbb{E}[P_{Trie}(w_{(i)}|w_\alpha)]\times \mathbb{E}[P_{Trie}(w_\alpha)]$ since both of the random variables on the RHS are independent by definition. Also since $\mathbb{E}[\hat{p}_\alpha]=p_{(1)}+p_{(2)}+\ldots +p_{(r)}$, we can use Eqn.(\ref{InductionEqn3})
and Eqn.(\ref{InductionEqn4}) to get that, for each $w_{(i)}$, $i=1,\ldots ,r$
\begin{align}
    \mathbb{E}[\hat{p}_{(i)}]&=\mathbb{E}[\hat{p}'_{(i)}]\times \mathbb{E}[\hat{p}_\alpha] \nonumber \\
    &=\frac{p_{(i)}}{p_{(1)}+p_{(2)}+\ldots +p_{(r)}} \times (p_{(1)}+p_{(2)}+\ldots +p_{(r)}) \nonumber \\
    &=p_{(i)} \nonumber
\end{align}
Hence for each $w_i \in T$, $\mathbb{E}[\hat{p}_i]=p_{i}$ when $|\mathcal{S}|=k+1$.\\
\qed\\
\\
While Theorem 3.1 justifies the use of Trie through equality of occurrence probability and the expected Trie probability at any stage of training, the next theorem provides a more desired result.
\\ \\
\textbf{Theorem 3.2:} For each $w_i\in\mathcal{S}$, $\hat{p}_i \xrightarrow{a.s.}p_i$
as $n\rightarrow \infty$.\\

\textit{Proof:} Let $\eta_j$ denote a node of the Trie such that each time any one of the words $w_{(1)},\ldots w_{(\eta_j)}$ from $\mathcal{S}$ is sampled and used for training, its $Counter$ variable is incremented by 1. Let $W_{\eta_j}=\{w_{(1)},\ldots w_{(\eta_j)}\}$.\\
Define, a random variable $\mathbbm{1}_n^{\eta_j}$ such that,
\[   
\mathbbm{1}_n^{\eta_j} = 
     \begin{cases}
       1 &\quad\text{if }n^{th}\text{ training word }\in W_{\eta_j}\\
       0 &\quad\text{otherwise.} \\ 
     \end{cases}
\]
For a fixed node $\eta_j$, $\mathbbm{1}_n^{\eta_j}$ are i.i.d random variables for $n\in \mathbbm{N}$. Also for each $n\in \mathbbm{N}$, $\mathbb{E}[\mathbbm{1}_n^{\eta_j}]=p_{(1)}+\ldots +p_{(\eta_j)}$. Note that the $Counter$ variable of $\eta_j$ is actually $\sum_{i=1}^{n}{\mathbbm{1}_i^{\eta_j}}$. Hence by using the Strong Law Of Large Numbers,
\begin{equation} \label{a.s.MainEqn}
    \frac{Counter(\eta_j,n)}{n}=\frac{\sum_{i=1}^{n}{\mathbbm{1}_i^{\eta_j}}}{n} \xrightarrow[n\rightarrow \infty]{a.s.} p_{(1)}+\ldots +p_{(\eta_j)}
\end{equation}
With the above in mind, we proceed for induction over the number of words in $\mathcal{S}$.\\
\begin{figure}
    \centering
    \includegraphics[width=4cm, height=4cm]{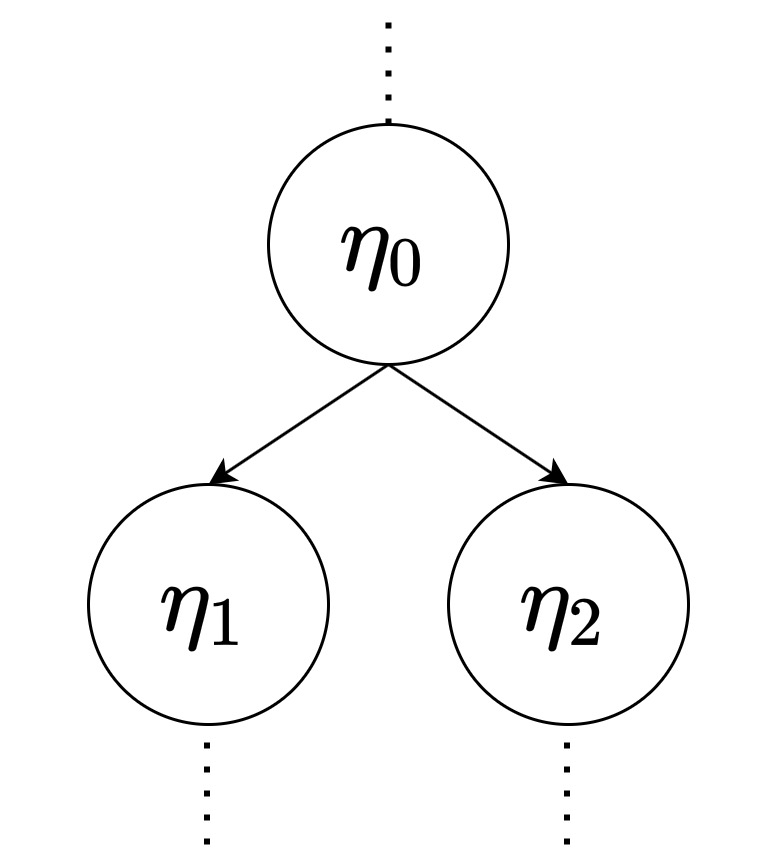}
    \caption{Base Case with Node Labels}
    \label{fig:Theorem2BaseCaseFig}
\end{figure}

\textbf{Base Case:} Let $\mathcal{S}$ contain two words $w_1$ and $w_2$ with occurrence probabilities $p_1$ and $p_2$. Then as in the previous theorem, we can get three cases which can be for our purposes be depicted by Fig.\ref{fig:Theorem2BaseCaseFig}. It is possible for $\eta_0$ to be the root note as well. It is also possible for exactly one of $\eta_1$ or $\eta_2$ to be a dummy node. This would cover all the three cases. \\
Now for $w_1,$
\begin{align}
    \hat{p}_1=&\frac{Counter(\eta_1,n)}{Counter(\eta_1,n)+Counter(\eta_2,n)} \nonumber \\
    =&\frac{Counter(\eta_1,n)/n}{Counter(\eta_1,n)/n+Counter(\eta_2,n)/n} \nonumber
\end{align}
Note that $Counter(\eta_1,n)/n \xrightarrow[n\rightarrow\infty]{a.s.}p_1$ and that $Counter(\eta_2,n)/n \xrightarrow[n\rightarrow\infty]{a.s.}p_2$. Hence, by using the Continuous Mapping Theorem \cite{van2000asymptotic}, we get that
\begin{equation}
    \hat{p}_1\xrightarrow[n\rightarrow\infty]{a.s.}\frac{p_1}{p_1+p_2}=p_1 \nonumber
\end{equation}
We can similarly show that $\hat{p}_2\xrightarrow[n\rightarrow\infty]{a.s.}p_2.$ Hence our base case i.e. $|\mathcal{S}|=2$ holds true.\\
The proof of the induction step is similar to that of Theorem 3.1 and is being skipped due to page constraint.\\
\qed\\\\
From Theorem 3.1 and Theorem 3.2 we conclude that from the perspective of Theory of Estimation and using the fact that almost sure convergence is stronger than convergence in probability, we get the following corollary\\
\\ \textbf{Corollary 3.3:} For each $w_i \in \mathcal{S}$, the Trie probability $\hat{p}_i$ is an unbiased and consistent estimator of the occurrence probability $p_i$.
\\ \\
Through the results above, we can infer that the new Trie can learn the occurrence probabilities for any set of words through sufficient training which in turn implies that the new Trie can adapt to the texting and typing style of any user when deployed on either a mobile phone, laptop or any such other device.
\begin{figure}
    \centering
    \includegraphics[width=\linewidth,height=7cm]{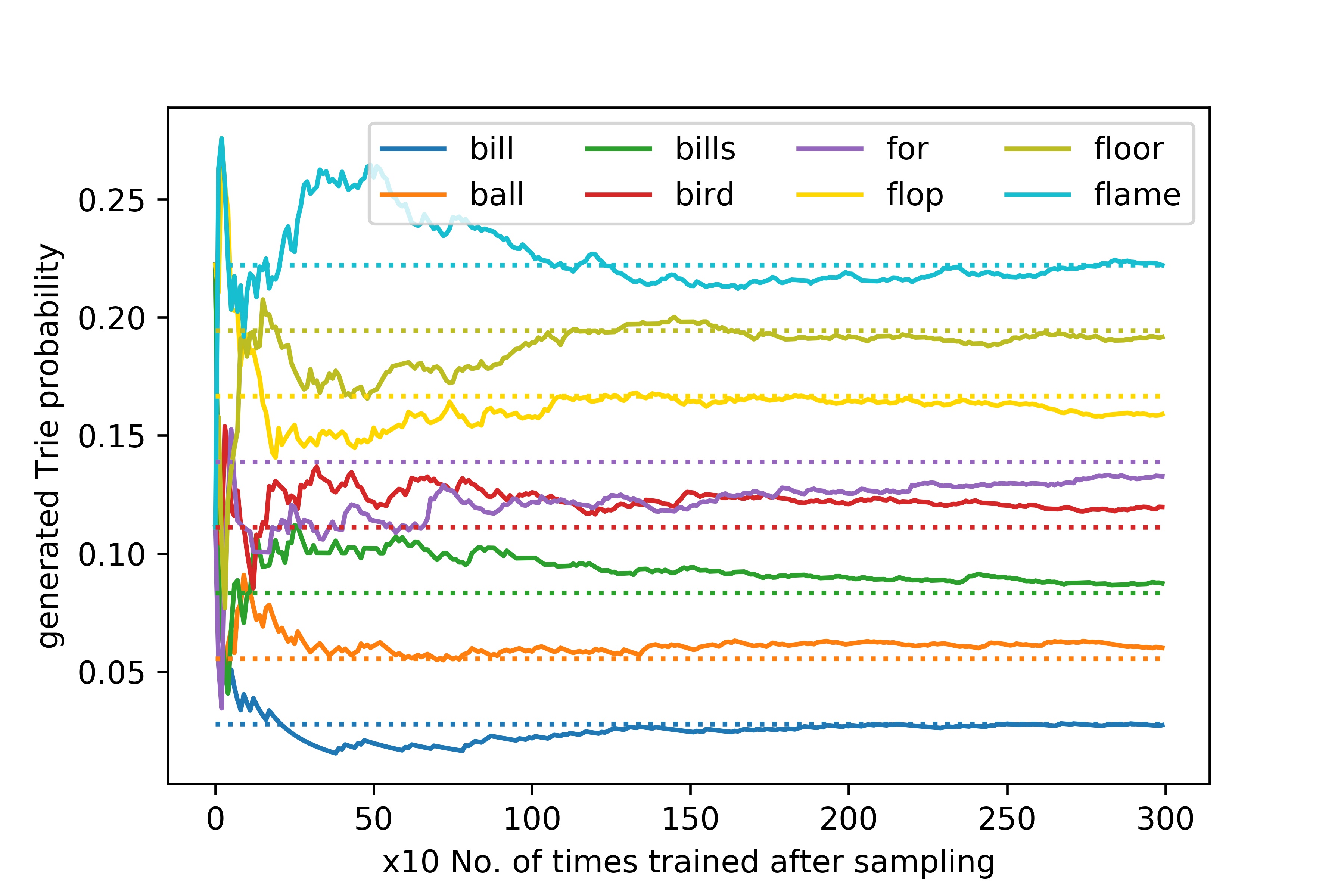}
    \caption{Almost sure convergence to occurrence probabilities}
    \label{fig:asfigure}
\end{figure}

\section{Error Checking Algorithm}\label{Section4}
In this section, we built upon the error correction schemes used in the baseline model.
\subsection{The Bayesian approach}
As mentioned earlier in Section \ref{Section2}, in \cite{inbook} the words in the suggestions list were ranked solely based on the Trie probabilities. However, following \cite{Keyboards2019ImprovingTA}, we use a Bayesian noisy channel approach to rank the words based on the type of error committed by the user. This requires assignment of probabilities to the types of error for which confusion matrices in \cite{inproceedings} can be used. Hence
\begin{equation}
     P(w|\Tilde{w})\propto P(\Tilde{w}|w)P(w) \nonumber
\end{equation}
where $w$ denotes a possible correction for $\Tilde{w}$. $P(\Tilde{w}|w)$ depends on the type of error committed to get $\Tilde{w}$ from $w$ and $P(w)$ is the occurrence probability of $w$ estimated using the Trie.

\subsection{Character Bigram}
It was observed in \cite{inbook} that the Trie could generate suggestions up to a Damerau-Levenshtein distance of one. This is because generating an exhaustive list of words at a DL distance of two is computationally very expensive and is of time complexity $\mathcal{O}(26^{2M})$ where $M$ is the length of the error word. However, 80\% of human errors lie within a DL distance of one and almost all within a DL distance two \cite{10.1145/363958.363994}. To partially overcome this, we propose a heuristic on the lines of beam search algorithm \cite{6591953}, wherein we maintain a character probability bigram denoted by $C$ where $C[i][j]$ represents the probability of the next letter being 'j' given that the current letter is 'i'. At each stage of error correction, we provide a score to each word say $`s_1s_2\ldots s_n'$ as:
\begin{equation}
    score(word)=\Pi_{i=1}^{n-1}{C[s_{i}][s_{i+1}]}
\end{equation}
and set a threshold, say $\Gamma$. If the score exceeds the threshold, the word is passed on to the next stage of error correction. This ensures that most of the intermediate low probability words are discarded while only the ones with a very high probability are passed on further.
\begin{figure}
    \centering
    \includegraphics[width=\linewidth,height=7cm]{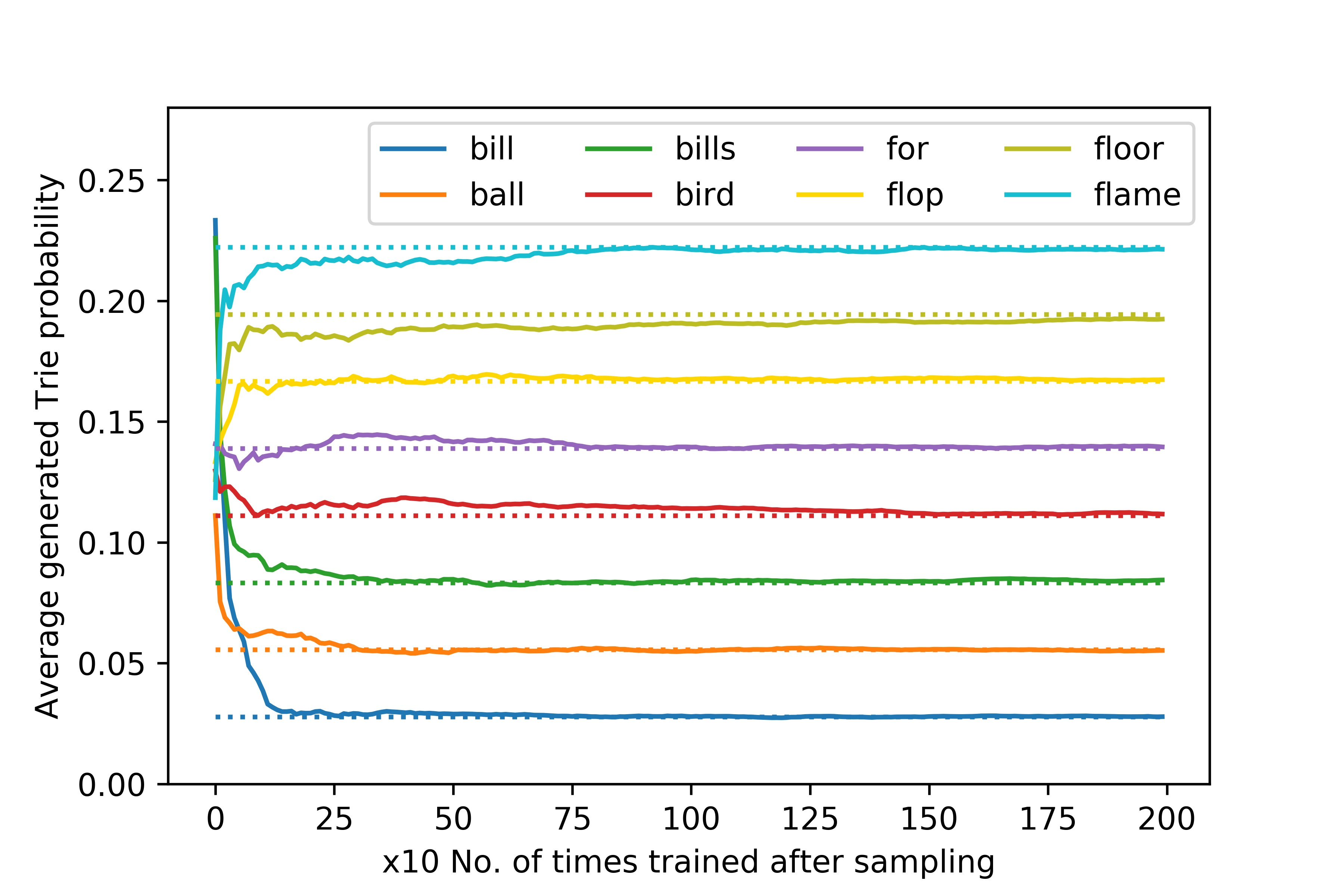}
    \caption{Equality of expectation to occurrence probabilites}
    \label{fig:efigure}
\end{figure}
\begin{figure*}
    \centering
    \includegraphics[width=\textwidth,height=10cm]{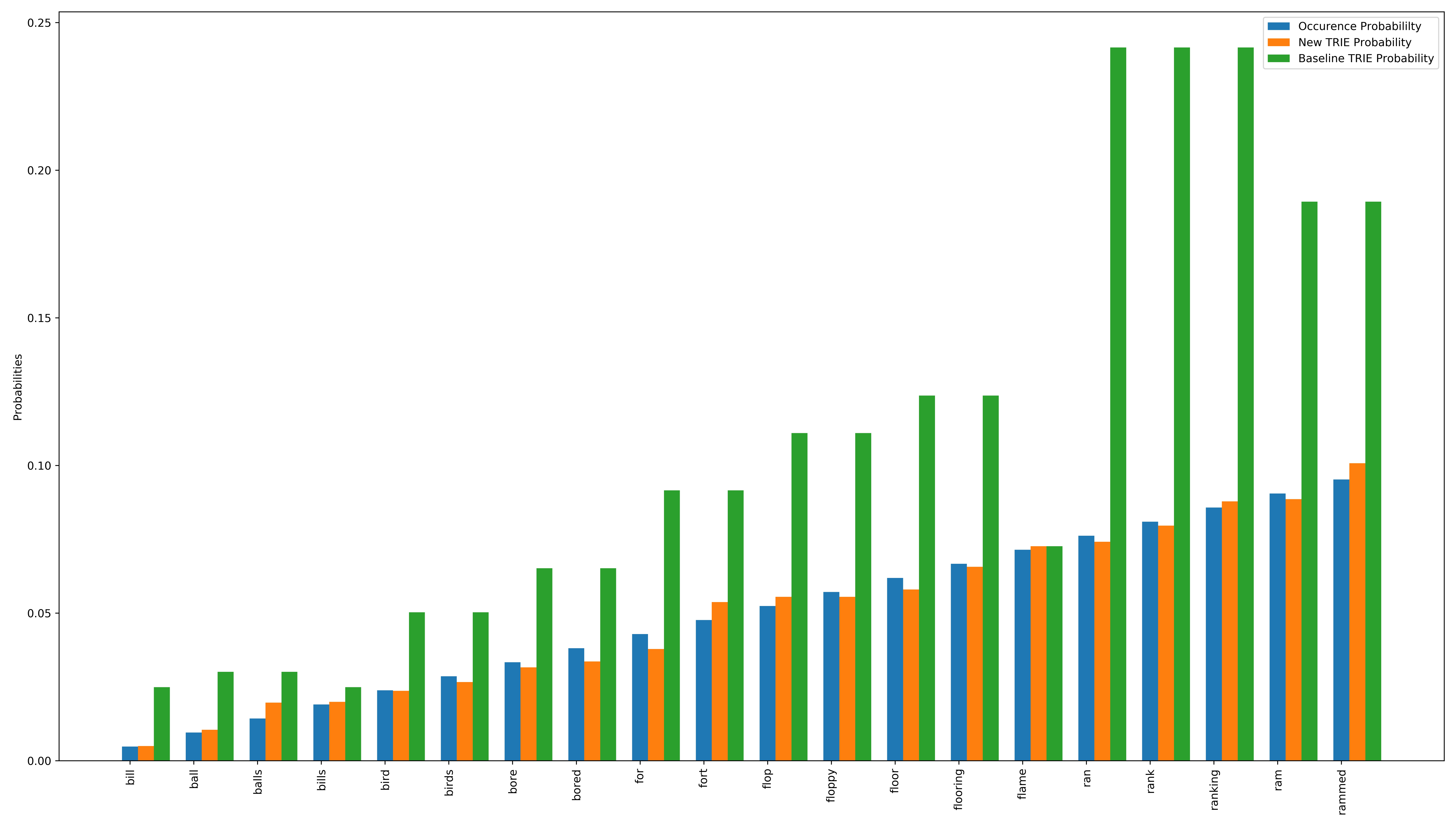}
    \caption{Performance comparison of the two TRIE based models}
    \label{fig:comparison}
\end{figure*}
\section{Simulations}\label{Section5}
\subsection{Almost sure convergence} \label{Section5.1}
In light of Theorem 3.2, a simulation was performed where we trained the new Trie using the training and probability generation algorithms defined in Section 3 with the eight words used in Section 2. The Trie probabilities denoted by the bold lines evidently converge to the occurrence probabilities denoted by the dotted lines in Fig.\ref{fig:asfigure}.
\subsection{Equality of expectation}
Similar to the simulation done in Section \ref{Section5.1}, we train 30 Tries and take the average of their probabilities. The bold lines can be seen to converge much faster to the dotted lines in Fig.\ref{fig:efigure} than in Fig.\ref{fig:asfigure}. This supports the theorem stating the equality of expectation of the Trie probabilities to the occurrence probabilities.
\subsection{Comparison}
In this simulation, we consecutively train both, the new Trie model proposed by us and the baseline Trie model using a corpus of twenty words. The assigned occurrence probabilities to these words are depicted in Fig.\ref{fig:comparison}. The new Trie clearly outperforms the one proposed baseline Trie. An important observation is that the baseline Trie probabilities clearly sum up to more than one, hence not a valid probability measure.

\subsection{Error correction}
We use a corpus of 3000 highest frequency English words\footnote{https://gist.github.com/h3xx/1976236} and use Zipf's Law\footnote{Value of the exponent characterising the distribution was set to 0.25} to fit\footnote{Not needed once deployed, model learns probability directly from the user.} a probability distribution over the ranked data. Comparative analysis with the baseline shown in Table \ref{ErrorCorrectionTable} clearly showcases superiority of the new model.
\begin{table}[h]
\caption{\label{tab:errorprobab}Comparison of output with baseline Trie}
 \centering
 \resizebox{\linewidth}{2.7cm}{
 \begin{tabular}{|c|c|c|c|c|} 
 \hline
 Impure & Target & Top 5 Suggestions &Rank &Rank in \cite{inbook}\\ [0.5ex] 
 \hline\hline
 tran & train & than, train , ran, trap, tan &2&5\\ 
 \hline
 lng & long & long, lang, log, leg, an, gang&1&4\\
 \hline
 aple&apple&pale, alle, able, apple, ample&4&5\\
 \hline
 beleive& believe &believe, believed, believes&1&1\\
 \hline
 gost& ghost & host, lost, most, past, ghost&5&1\\
 \hline
 moble & noble & noble, nobler, nobles&1&2\\
 \hline
 cuz & cause & use, case, cause, cut, cup&3&8\\
 \hline
 cin&seen &in, can, sin, son, skin &13&17\\
 \hline
 dem&them&them, then, den, chem, the&1&11\\
 \hline
 m8&mate&mate, might, eight, ate, mare&1&6\\
 \hline
 thx&thanks&the, thy, tax, thanks, them&4&1\\
 \hline
 h8&hate&hate, height, hare, ate, haste&1&-\\
 \hline
 \end{tabular}}
 \label{ErrorCorrectionTable}
\end{table}

\section{Conclusions}\label{Section6}
In the paper, we first pointed out a a limitation in the Trie based probability generating model proposed in existing literature, to overcome which, we proposed a structural modification, a training algorithm and a probability generating scheme. We further proved rigorously that the new Trie generated probabilities are an unbiased and consistent estimator of the occurrence probabilities. These occurrence probabilities vary user to user which the new Trie is capable of adapting to. We performed simulations, the results of which strongly support both the presented theorems and demonstrated superiority in error
correction.

\bibliography{refs}
\end{document}